\documentclass{article}
\usepackage{spconf,amsmath,graphicx}
\usepackage{xspace}
\usepackage{times}
\usepackage{epsfig}
\usepackage{amssymb}
\usepackage{ bbold }
\usepackage{bbm}

\makeatletter
\DeclareRobustCommand\onedot{\futurelet\@let@token\@onedot}
\def\@onedot{\ifx\@let@token.\else.\null\fi\xspace}

\def\eg{\emph{e.g}\onedot} 
\def\ie{\emph{i.e}\onedot}

\def\etal{\emph{et al}\onedot}
\makeatother

\newcommand{\squeezeupSmall}{\vspace{-2mm}}
\newcommand{\squeezeup}{\vspace{-4mm}}
\usepackage{hyperref}

\title{Compositional Sketch Search}
\name{Alexander Black$^{\star}$ \qquad Tu Bui $^{\star}$ \qquad Long Mai$^{\dagger}$ \qquad Hailin  Jin$^{\dagger}$ \qquad John Collomosse$^{\star \dagger}$}
  
\address{$^{\star}$ CVSSP, University of Surrey \qquad $^{\dagger}$ Adobe Research}
\begin{document}
%
\maketitle
\begin{abstract}
We present an algorithm for searching image collections using free-hand sketches that describe the appearance and relative positions of multiple objects\footnote{\url{https://github.com/AlexBlck/compsketch}}.  Sketch based image retrieval (SBIR) methods predominantly match queries containing a single, dominant object invariant to its position within an image.  Our work exploits drawings as a concise and intuitive representation for specifying entire scene compositions. We train a convolutional neural network (CNN) to encode masked visual features from sketched objects, pooling these into a spatial descriptor encoding the spatial relationships and appearances of objects in the composition.  Training the CNN backbone as a Siamese network under triplet loss yields a metric search embedding for measuring compositional similarity which may be efficiently leveraged for visual search by applying product quantization.
\end{abstract}
\begin{keywords}
Sketch,  Visual Search, Composition.
\end{keywords}
\section{Introduction}
\label{sec:intro}

Sketches are a concise and intuitive way to  visually describe the composition of a scene \ie the {\em appearance} and the {\em relative spatial arrangement} of  objects present.  Significant progress has been made in harnessing hand-drawn sketched  queries to drive large-scale visual search of image \cite{Cao2010,sketchstyle2017} and video \cite{Collomosse2009,hu2013markov} collections.  Yet existing sketch based image  retrieval (SBIR) algorithms typically ignore composition,  matching only a single sketched object irrespective of its position on the query canvas.  Moreover, training often assumes sketched objects to match a single dominant object occupying the majority of the image -- complicating partial image matching and retrieval of smaller objects present.

This paper contributes a technique for {\em compositional sketch search}; to search images by matching a query sketch containing multiple objects taking into account both their appearance and their relative position on the sketch canvas.  Our core technical contribution is a method for encoding visual features from a sketch encoder into a spatial map, forming a latent space (`search embedding') for measuring content similarity.  We train our encoder in a contrastive architecture to encourage learning of a metric  search embedding, using object compositions sampled from OpenImages \cite {Kuznetsova2020}.

\begin{figure}[t!]
    \centering
    \includegraphics[width=\linewidth,height=3.5cm]{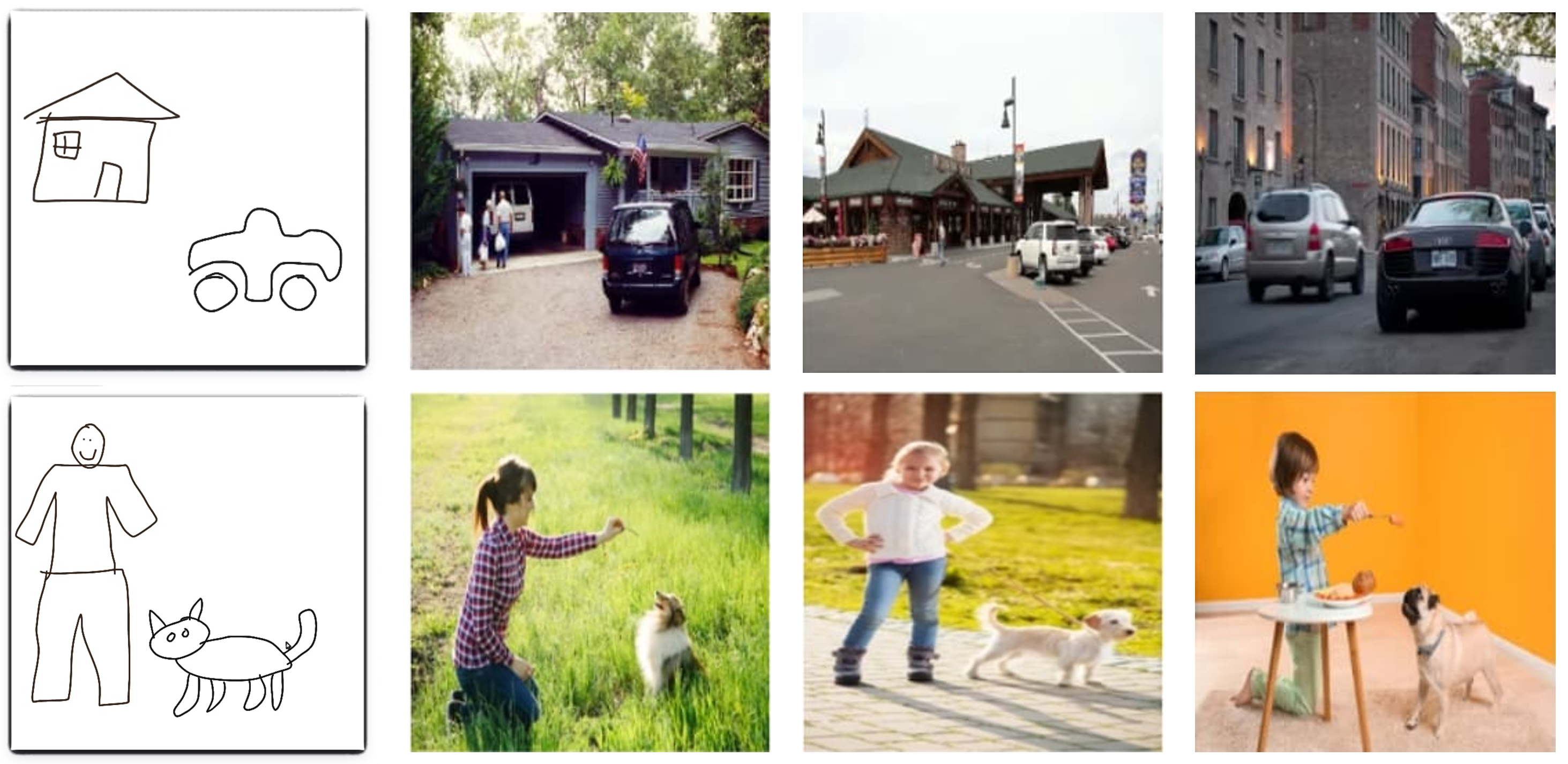}
    \caption{Sketched compositions (queries) comprising multiple objects and the corresponding results from OpenImages \cite{Kuznetsova2020} depicting those objects in similar spatial arrangements. }
    \label{fig:teaser}
	\squeezeup
\end{figure}

\section{Related Work}
\label{sec:lit}

Early dictionary learning SBIR methods leveraged wavelet \cite{Sciascio1999},  edge-let \cite{Cao2010}, key-shape \cite{Saavedra2015} and sparse gradient features \cite{Eitz2011,Hu2013} to match sketches to edge structures in images.  Deep learning approaches extensively apply convnets  for cross-modal representation learning; exploring joint search embeddings for matching structure between sketches and images.  Early approaches learned mappings between edge maps and sketches using contrastive \cite{wang2015sketch3d}, siamese \cite{qi2016} and triplet networks~\cite{Bui2017compact}. Fine-grained SBIR was explored by Yu \etal \cite{yu2016SketchMeShoe} and Sangkloy \etal \cite{sangkloy2016sketchy} who used a three-branch CNN with triplet loss \cite{gordo2016deep}.  Bui \etal learned a cross-domain embedding through triplet loss and partial weight sharing between sketch-image encoders~\cite{Cag} to yield state of the art results. Stroke sequence models have also been explored to learn search embeddings \cite{Xu2018sketch,Ribeiro2020Sketchformer}. With the exception of very recent work \cite{Liu2020SceneSketcher} all the above approaches require single object sketches rather than scenes with multiple objects.  Liu \etal~\cite{Liu2020SceneSketcher} encode a scene graph, requiring explicit pre-detection of objects. By contrast our approach requires no explicit detection step to index images, instead building upon spatial visual search for photos \cite{conceptcanvas} to match sketched object layout.

\begin{figure*}[t!]
    \centering
    \includegraphics[width=1.0\linewidth,height=4.5cm]{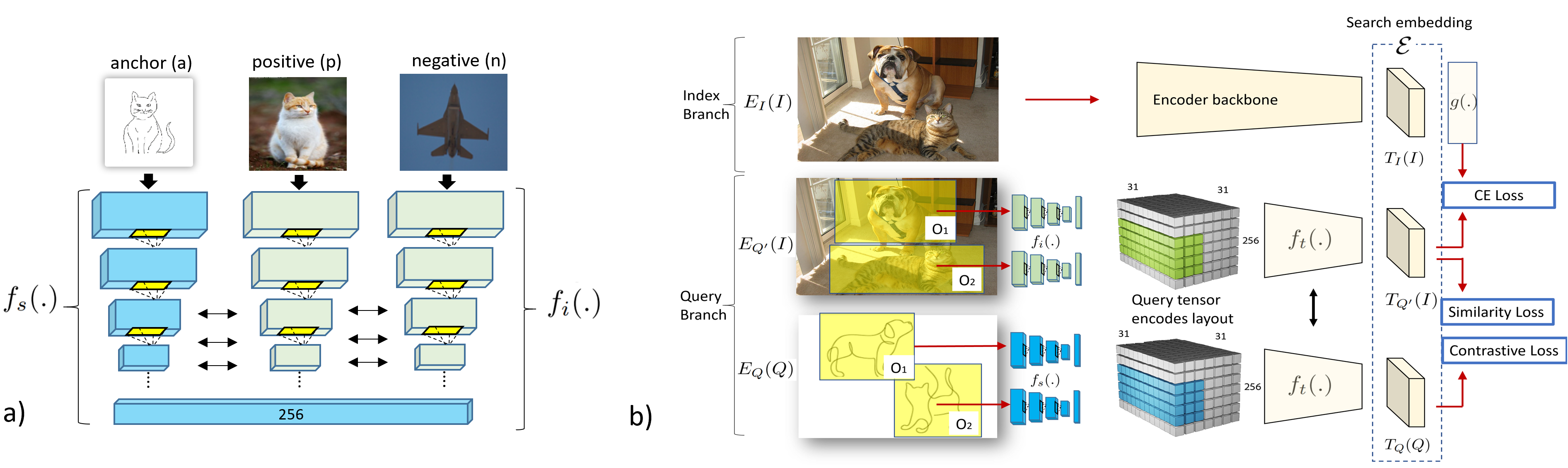}
    \caption{Proposed architecture.  (a) Our approach builds upon  \cite{Cag} that maps a single-object sketch (blue, $f_s(.)$) and image (green, $f_i(.)$) to a common embedding (black arrows show shared weights).  (b) We encode multiple objects $O_i$ into a query tensor that approximates object layout with a similar spatial layout of appearance vectors.  A spatial encoder  $f_t(.)$ is learned to map the query tensor into our search embedding $\mathcal{E}$.  During training, the query tensor is formed using images ($E_{Q'}(I)$) and at query time using sketches ($E_{Q}(Q)$). Branch ($E_I(I)$)  indexes the search corpus via an encoder backbone (\eg GoogLeNet).}
    \label{fig:arch}
\end{figure*}

\squeezeup
\section{Spatial Sketch-based Image Retrieval}
\label{sec:ssbir}
\squeezeupSmall
We propose a method for spatial-aware sketch based image retrieval (SSBIR) that accepts a raster query ($Q$) containing a free-hand sketched composition that describes the appearance and relative positions of potentially many objects $O=\{O_1,...,O_n\}$. We learn a joint search embedding $\mathcal{E}$ into which sketches $E_Q(Q) \mapsto \mathcal{E}$ and images $\mathcal{I}=[I_1,...,I_m]$ indexed by the search $E_I(I) \mapsto \mathcal{E}$ may be mapped via learned encoders $E_Q(.)$ and $E_I(.)$ for the sketch and image domain respectively.  The  $L_2$ distance $|E_Q(Q)-E_I(I)|_2$ ranks images in the search corpus $\mathcal{I}$  by similarity to sketch $Q$.

\squeezeupSmall
\subsection{Network Architecture}

Our method extends the state of the art single-object multi-stage SBIR method of Bui \etal \cite{Cag} (hereafter mSBIR), which leverages a triplet architecture with GoogLeNet backbone and partially-shared weights (in late layers) to encode both sketches and images to a common feature embedding (Fig.\ref{fig:arch}a). We refer to these encoding functions as $f_s(.)$ and $f_i(.)$ respectively. 
mSBIR learns $f_s(.)$ and $f_i(.)$ via a triplet network comprising an anchor (a) branch accepting a sketch query and positive/negative (p/n) branches that accept an image as input.  The common feature embedding is read out from a $C=256$ channel fully connected (fc) layer shared across all network branches.  mSBIR forms triplets using single-object sketches from the TU-Berlin dataset \cite{Tuberlin}, accompanied by a pair of images containing an object of the same (p) and different (n) class.  


We build upon mSBIR to tackle sketched compositions by independently encoding each object $O_i$, and aggregating the resulting features into a query tensor $T_Q(Q)$.  Given a raster $Q$ of resolution $W \times H$ pixels, let $R[Q,O_i]$ be a  cropping operator yielding sub-image of $Q$ delimited by the bounding box of $O_i$. Let $\mathbbm{1}(O_i)$ be a $W \times H$ field of scalar weight $1/\kappa$, where $\kappa \in [1,n]$ counts the number of objects overlapping each pixel. Let $[\mathbbm{1}]_{\times C}$ duplicate that field across $C$ channels. The mask is The $1 \times C \times W \times H$ query tensor is formed by aggregating the feature embeddings of all $n$ objects  in $Q$.
\begin{equation}
T_Q(Q) = \mathrm{MP}_{N \times N} \left[ \sum_{i=1}^n f_s(R[Q,O_i]) \odot  [\mathbbm{1}]_{\times C} (O_i) \right]
\squeezeupSmall
\end{equation}
where $\odot$ indicates in-place multiplication, and $MP_{N \times N}[.]$ is a maxpooling operator that downsamples the tensor resolution to $C \times N \times N$, we use $N=31$ for our experiments.  The resulting tensor comprises a zero vector for each pixel position unoccupied by an object, otherwise the vector average  pools features of objects  that overlap that pixel position (overlapping objects are permitted).


$T_Q(Q)$ encodes sketch $Q$, however a similar process be applied to compute a tensor $T_{Q'}(I)$ from an  image $I$ containing potentially many objects, by leveraging feature encoder $f_i(.)$.  This is used during training only (subsec.~\ref{sec:train}). 
\begin{equation}
T_{Q'}(I) = \mathrm{MP}_{N \times N} \left[ \sum_{i=1}^n f_i(R[I,O_i]) \odot  [\mathbbm{1}]_{\times C} (O_i) \right]
\end{equation}

We next define $f_t(.)$, a spatial feature encoder that accepts input tensor $T$ returned by either the encoding function for query sketches ($T_Q(Q)$) or  training images ($T_{Q'}(I)$). We model the function  $f_t(.)$  via a convnet with three convolution layers with $3 \times 3$ kernel size, interleaved by two max-pooling layers of stride 2 each followed by batch normalisation and ReLU activation (Fig.\ref{fig:arch}b).  The purpose of $f_t(.)$ is to map the tensor representation into $\mathcal{E}$ thus enabling both sketches and training images passed down the query branch of our network to be encoded to the search embedding, via functions $E_Q(Q)$ and $E_{Q'}(I)$ respectively. 
\begin{eqnarray}
\label{eq:querytensor}
E_Q(Q)&=&f_t^*(T_Q(Q))\\ \nonumber
E_{Q'}(I)&=&f_t^*(T_{Q'}(I))
\end{eqnarray}
The output of $f_t(.)$ a tensor. Therefore to map $I$ or $Q$ to $\mathcal{E}$ via the query branch,  the output is flattened (indicated as $f_t^*(.)$).

\squeezeupSmall
\subsection{Image Indexing Branch}

To index images $I \subset \mathcal{I}$ with the search corpus, we incorporate an 'indexing' branch in our network -- adapting convnet backbones such as GoogLeNet or ResNet. Each image $I$ is passed through the early convolution stages of the network, to yield a  $7 \times 7 \times D$ tensor $T_I(I)$ (so matching of the dimension of query-derived tensor  $T_Q(.)$, eq. \ref{eq:querytensor}).  For example, if using GoogLeNet to learn $T_I(I)$, we use the first five convolutional stages (to pool5/5) of the network, where $D=832$.  Images are encoded by flattening the resulting tensor:
\begin{equation}
E_I(I)=T_I^*(I)
\end{equation}

\squeezeupSmall
\subsection{Learning the Query Encoders}
\label{sec:train}

The image-derived query tensor encoder $T_{Q'}(I)$ is used to learn the spatial feature encoder  $f_t(.)$, whilst simultaneously fine-tuning the indexing branch of the network  $T_I(I)$.  The single-object common embedding, via $f_s(.), f_i(.)$, is trained offline beforehand.  We initialize $T_I(I)$ weights  from a pre-trained ImageNet classifier \cite{imagenet}.

The OpenImages \cite{Kuznetsova2020} dataset provides images and associated bounding box annotations for scene objects.  Rather than training with  paired sketch-image data, which is unavailable in volume for compositions, we exploit the common single-object sketch embedding via $f_s(.)$ and $f_i(.)$. Random pairs of photographic images $(I,I_{\mathrm{neg}})$ are sampled from the OI  training set (c.f. subsec.~\ref{sec:dataset}).
We build upon the spatial visual search approach of Mai \etal \cite{conceptcanvas} to learn $f_t(.)$ and $E_{Q'}(I)$ via three loss terms.  First, a similarity loss encourages $E_{Q'}(.)$ and $E_I(.)$ to map to a common embedding:
\begin{equation}
L_\mathrm{sim}= 1 - \cos (E_I(I),E_{Q'}(I))
\end{equation}

Second, a discriminative loss is added via a 4096-D fc layer $g(T_I(.))$ to the indexing branch for training purposes, and minimizing Cross-Entropy (CE) loss:
\begin{equation}
L_\mathrm{ce}= \mathrm{CE}(g(T_I(I),c(I)))
\end{equation}
where $c(I)$ is a class likelihood vector \ie non-zero for the classes present in the image.  Third, a contrastive loss computed between the encoding of a random 'irrelevant' image $T_I(I_{\mathrm{neg}})$ and encodings of image $I$ passed down the query ($T_Q(I)$) and the indexing ($T_I(I)$) branches.

\begin{figure}[t!]
    \centering
    \includegraphics[width=0.9\linewidth,height=7cm]{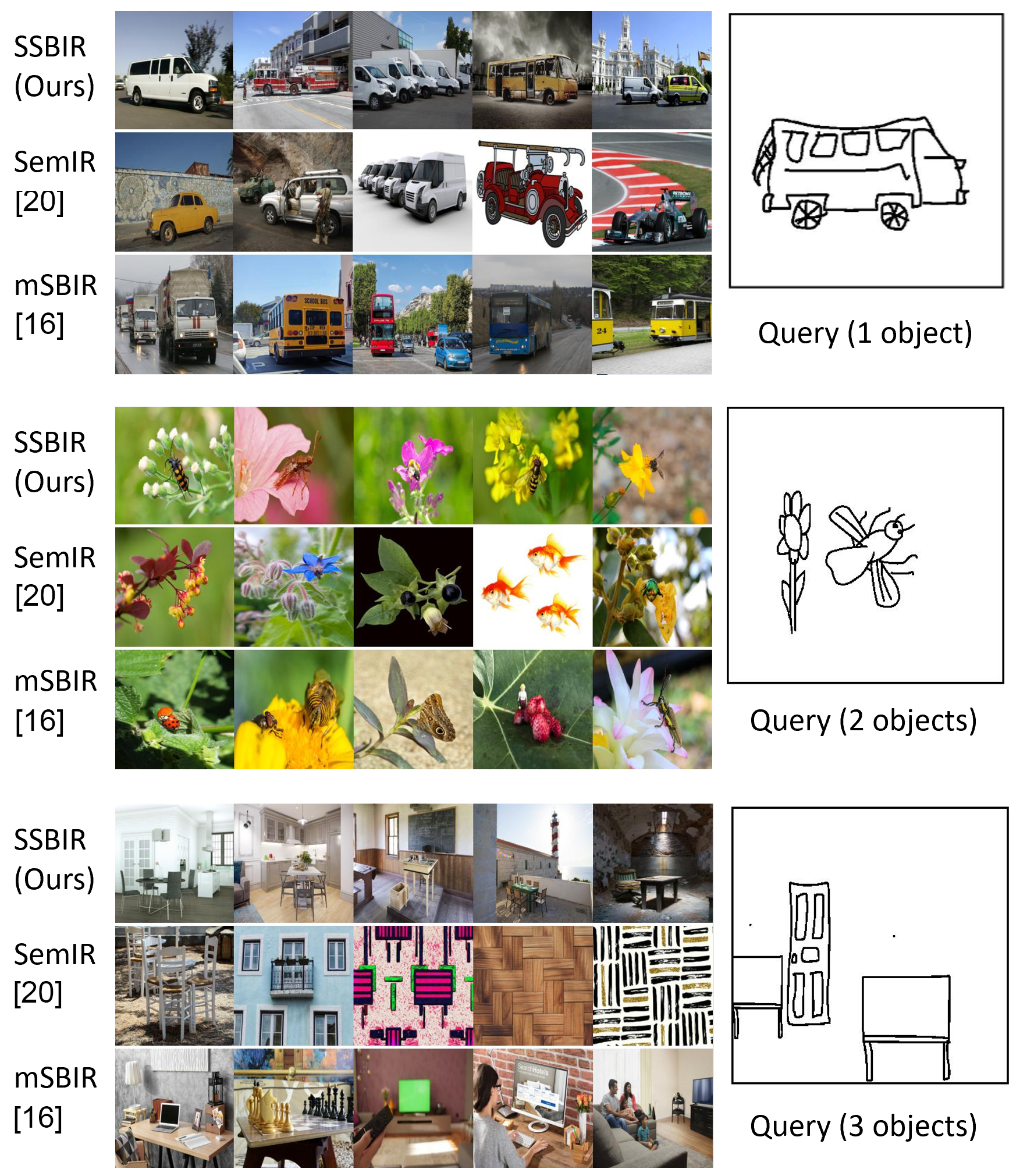}
    \caption{Representative top-5 results querying 1-, 2- and 3-object compositional sketches on Stock4.5M. }
    \label{fig:eg}
    \vspace{-1em}
\end{figure}

\begin{eqnarray}
\label{eq:totalloss}
\mathcal{L}_\mathrm{con}(I,I_{\mathrm{neg}}) = &[ m + cos(E_{Q'}(I) - E_I(I)) - \nonumber \\
 &cos(E_{Q'}(I)-E_I(I_{\mathrm{neg}})) ]_+
\end{eqnarray}
where $m=0.3$ is a margin promoting convergence, and $[x]_+$ is the non-negative part of $x$.  The total loss is a weighted sum (80\%, 15\%, 5\%) of these three terms respectively.

\squeezeupSmall
\subsection{Compact Representation of $\mathcal{E}$}
\squeezeupSmall
The high dimensionality of $\mathcal{E}$ makes it infeasible for SSBIR to scale search to large collections \eg $>1M$. To address this we project and binarize $\mathcal{E}$ via 2-step Product Quantization (PQ) \cite{johnson2019billion}:
\begin{equation}
    \mathcal{B} = q_1(\mathcal{E}) + q_2(\mathcal{E} - q_1(\mathcal{E}))
\end{equation}
where $q_1$ is a coarse quantizer using KMeans and $q_2$ is the fine-level PQ applying on the residual data after $q_1$. We empirically set 1600 as number of KMeans clusters for $q_1$ and $16$ bytes for the PQ hash code output of $q_2$ (see sup-mat. for analysis on hash code length). We linearly project $\mathcal{E}$ to a lower dimensional space as a pre-process step prior to PQ, as suggested in OPQ \cite{ge2013optimized}.



\begin{table}[t!]
\centering
\begin{tabular}{lccc}
Method          & mAP & NDCG \\ \hline
SSBIR (Proposed)   &  \textbf{0.260}   &  \textbf{0.534}  \\
SemIR~\cite{conceptcanvas}   &  0.188  &  0.412  \\
mSBIR~\cite{Cag}   &  0.164   &  0.384 \\ \hline
SSBIR-GoogleNet \cite{Googlenet2015} &  \textbf{0.260}   &   \textbf{0.486}   \\
SSBIR-VGG11 \cite{vgg} &  0.198   &   0.450   \\
SSBIR-ResNet50 \cite{He2015}      &   0.058  & 0.340 \\    
SSBIR-MobileNet-V2 \cite{mobilenetv2}  &   0.187  &  0.414 \\   
SSBIR-EfficientNet-B0 \cite{efficientnet} &   0.183  & 0.434\\ \hline
\end{tabular}
\caption{Performance of the proposed method versus baselines \cite{conceptcanvas,Cag} (upper), and versus variants of the proposed method with different backbones (lower).}
\label{tab:results}
\vspace{-1em}
\end{table}

\begin{figure*}[t!]
    \centering
        \includegraphics[width=0.3\linewidth,height=3cm]{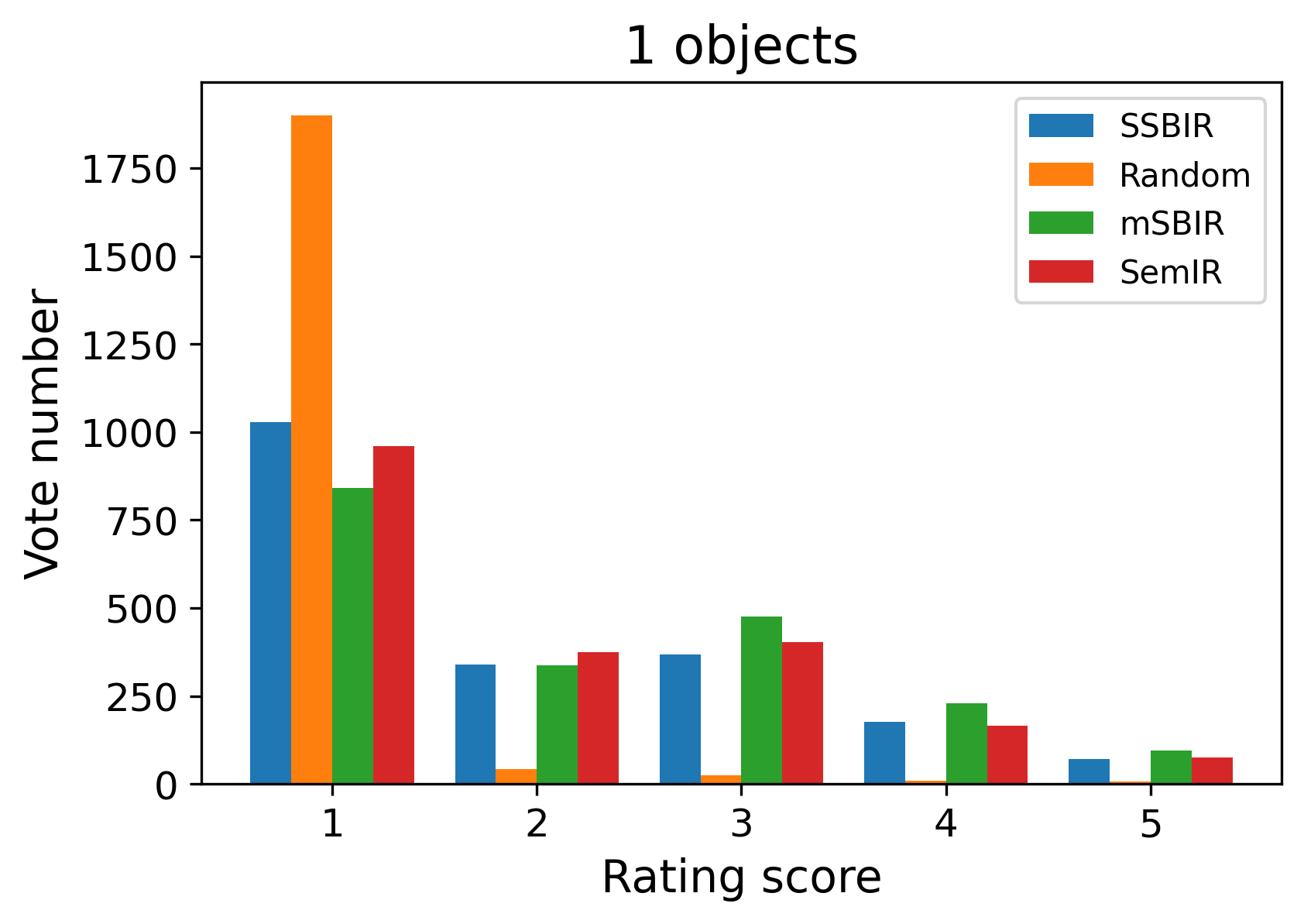}
        \includegraphics[width=0.3\linewidth,height=3cm]{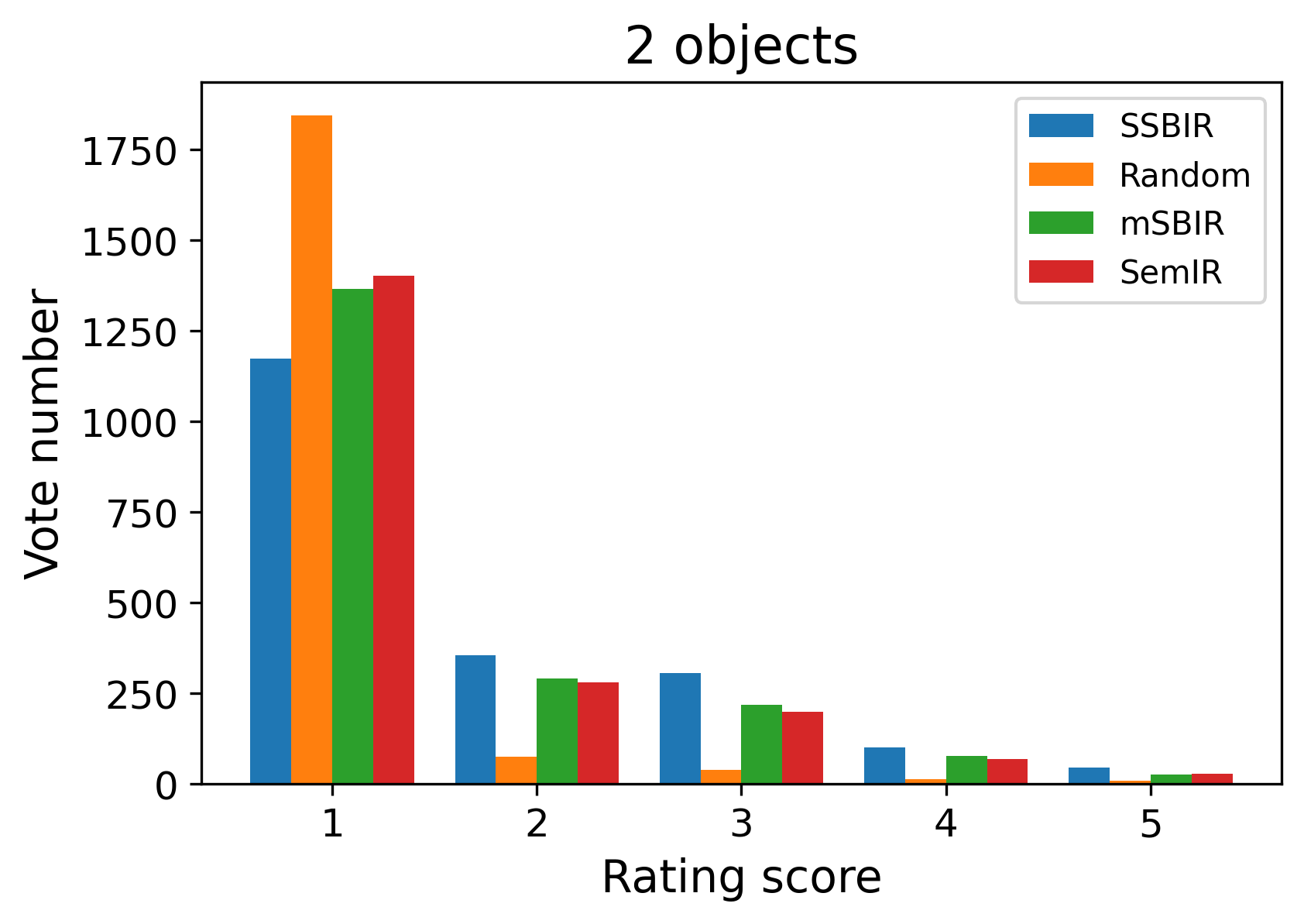}
        \includegraphics[width=0.3\linewidth,height=3cm]{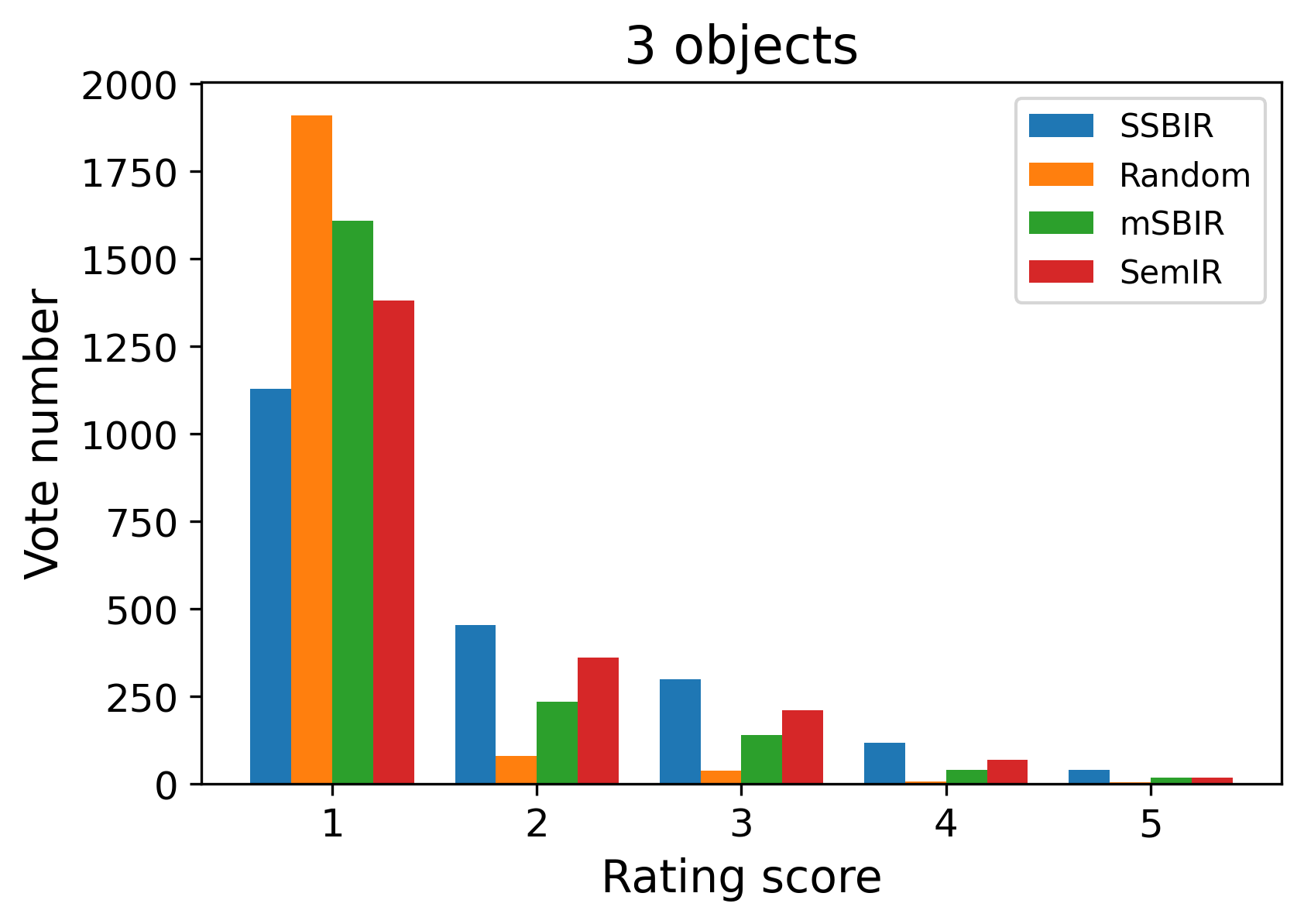}
    \caption{Amazon Mechanical Turk (MTurk) user study on the Stock4.5M dataset, for 1 (left), 2 (middle) and 3 (right) object queries. Each histogram shows the score distribution (1=poor, 5=good) for each method and a control (random) response.}
    \label{fig:turk}
\end{figure*}

\section{Experiments and Discussion}
\label{sec:exp}
\vspace{-1em}
We evaluate the performance of our compositional SBIR method, contrasting performance against single \cite{Cag} and multiple object \cite{conceptcanvas} baselines. We trained and tested all models on a 12GB GTX Titan-X GPU using the ADAM optimizer with learning rate 0.0001 and weight decay $5e-4$. 
\squeezeupSmall
\subsection{Datasets}
\label{sec:dataset}
\indent \textbf{OI-TrainVal} is the largest public dataset with object-level annotations \cite{Kuznetsova2020}; we use version 6 with $\sim$2M images of 801 classes and 16M bounding box annotations.  We use the public training/validation partitions to train our model.  We remove images that do not have class overlap with TU-Berlin, as well as object classes that are too broad (\eg mammal, furniture,...). The final set (hereafter, OI-TrainVal)  has 1.3M training and 26K validation images of 141 classes. 

\textbf{OI-Test-LQ} consists of 125k test images, obtained from the public OpenImages (OI) v6 test set via the class filtering applied for OI-TrainVal.  In addition, we synthesise 11K sketched compositions that serve as the query set for evaluation.  To construct the sketch set, we randomly sample 11K images (and their associated bounding boxes) from the OI test set.  Single object sketches from the TU Berlin dataset \cite{Tuberlin} that match the bounding box classes are positioned on a single canvas, to create the composition. We also sample a smaller set of 900 queries from the 11K set (hereafter, OI-Test-SQ) for our peripheral studies. 

\textbf{Stock4.5M} is a 4.5 million unwatermarked image dataset collected from \textbf{https://stock.adobe.com}.  A query set of  100 sketches is constructed in a similar way to the OI-Test-SQ query set, balancing the count of objects present (\ie the number of sketches containing 1, 2 and 3 objects are 33, 33, and 34 respectively).   Stock4.5M is for evaluation only, to test  scalability and domain generalization beyond OI-TrainVal.

\squeezeupSmall
\subsection{Evaluation metrics}
We evaluate ability to retrieve images that match the sketched object queries in terms of both semantic categories and spatial layout of objects. Our relevance score between query sketch $Q$ and database image $I$ is defined as:
\begin{equation}
    \mathcal{R}(Q,I) = \frac{1}{\left| B_Q\right|}\sum_{b_i\in B_Q}{\max_{b_j\in B_I}{\mathbb{1}_{[c(b_i)=c(b_j)]}\frac{b_i\cap b_j}{b_i\cup b_j}}}
    \label{eq:rel}
\end{equation}
where $B_Q$ and $B_I$ are sets of object bounding boxes in $Q$ and $I$, and where $\mathbb{1}_{[c(b_i)=c(b_j)]} \in \{1,0\}$ evaluates to 1 if the classes of the objects inside bounding boxes $b_i$ and $b_j$ match. $\mathcal{R}(Q,I)$ is continuous in range $[0,1]$ and can be binarized via  threshold $\tau$ - $\overline{\mathcal{R}}_\tau(Q,I) = 1$ if $\mathcal{R}(Q,I)>\tau$ otherwise 0.

Given this relevance score, we evaluate: (i) Mean Average Precision (mAP); (ii) Normalized Discounted Cumulative Gain (NDCG) which down-weights lower-ranked images.  mAP requires binary relevance, so we define $\overline{\mathcal{R}}_\tau(Q,I) = 1$ if $\mathcal{R}(Q,I)>\tau$ otherwise 0, with $\tau$ a threshold value (see sup-mat for analysis). NDCG works on continuous scores, $\overline{\mathcal{R}}_\tau(Q,I) = \mathcal{R}(Q,I)$ if $\mathcal{R}(Q,I)>\tau$ otherwise 0. We compute mAP and NDCG for the top $200$ results.

\begin{table}[t!]
\centering
\begin{tabular}{lccc}
Method                       & P@20      & mAP    & NDCG   \\ \hline
SSBIR (proposed)             & \textbf{0.432} & \textbf{0.323} & \textbf{0.737} \\
SemIR~\cite{conceptcanvas}    & 0.345 & 0.237 & 0.687 \\
mSBIR~\cite{Cag}             & 0.344 & 0.245 & 0.677 \\
Random*                      & 0.038 & 0.009 & 0.467
\end{tabular}
\caption{MTurk user study on top-20 retrieved results for the Stock4.5M image dataset, using 100 queries. * indicates a control set created by returning images at random.}
\label{tab:mturk}
\end{table}

\squeezeupSmall
\subsection{Evaluating baselines and architectures}
\label{sec:eval_arch}
We compare the proposed methods with two other baselines: \textbf{mSBIR} -- the state-of-art single object multi-stage SBIR \cite{Cag}, and \textbf{SemIR} -- the spatial semantic image retrieval method \cite{conceptcanvas}. Since SemIR only accepts keyword queries, we convert the sketched objects into category names whilst retaining its bounding box positions. Unless otherwise specified, $\tau$ is set to 0.5 for mAP and NDCG (see sup-mat. for study of $\tau$).

Table~\ref{tab:results} (top) reports the mAP and NDCG performance of our proposed method and the baselines on OI-Test-LQ. SSBIR outperforms the closest competitor by a large margin (by 7\% on mAP and 12\% on NDCG). This is significant because SSBIR embedding also encodes sketch appearance rather than just the word2vec embedding of class names as in SemIR \cite{conceptcanvas}. mSBIR \cite{Cag} performs  worst as it cannot encode spatial layout nor images with multiple objects.

We studied the effect of different CNN backbones for the image indexing branch $E_I(.)$ using OI-Test-SQ (Table~\ref{tab:results}, bottom). The GoogleNet backbone \cite{Googlenet2015} outperforms others \eg ResNet \cite{He2015} and EfficientNet \cite{efficientnet}. This may be due to a match with the  
\cite{Cag} use of GoogleNet within the single-object common embedding. All comparisons are statistically `very significant' (t-test; $\rho \ll 0.01$) except for GoogleNet versus VGG11 ($\rho = 0.148$). We therefore adopt GoogleNet.

\squeezeupSmall
\subsection{Evaluating large-scale retrieval}
\label{sub:mturk}


We conduct an user study on the Stock4.5M dataset using  Amazon Mechanical Turk (MTurk). We retrieve the 20 top-ranked  images for each of 3 methods: our SSBIR, and baselines SemIR \cite{conceptcanvas} and mSBIR \cite{Cag}. For a given query, we group the returned images of the same rank across these methods along with a random image to form an annotation task. We then ask 3 annotators to score each of the images in terms of its semantic and spatial relevance to the sketched query (scale: 1 means `completely irrelevant' and 5 means `correct objects in the exactly same pose and location'). 

All methods are \emph{significantly} better than random (t-test; $\rho<< 0.05$). Fig.~\ref{fig:turk} shows for 1-object queries, the performance of SSBIR is on par with SemIR~\cite{conceptcanvas} ($\rho=0.175$ indicates no significance in the rating distributions of these two methods) and slightly lower than mSBIR. For multi-object queries SSBIR receives \emph{significantly} more 3-5 ratings than both baselines. Tab.~\ref{tab:mturk} averages  the user ratings of each query-image pair yielding a relevance score $\mathcal{R}$ equivalent to eq.~\ref{eq:rel}.  We threshold the relevance score at 2 to compute mAP and NDCG metrics. We also report Precision at rank $k=20$, for all $20$ images annotated.  SSBIR outperforms baselines on all 3 metrics by a large margin. Several retrieval examples are given in Fig.~\ref{fig:eg}. mSBIR often works best on single-object sketches, whilst SemIR takes only semantic information into account and disregards appearance.

\section{Conclusion}
\label{sec:con}
We proposed a SBIR method for searching image collections using {\em compositional} sketches containing multiple objects.  Visual features from scene objects are encoded into a spatial feature map that is compressed via product quantization (PQ).    Our approach requires no object detection step. We show that explicitly training for multiple object SBIR yields statistically significant performance gains over single-object SBIR, and improved appearance recall using sketches versus labelled bounding boxes.  



\bibliographystyle{IEEEbib.bst}
\bibliography{refs.bib}

\end{document}